\documentclass[english]{article} 
\usepackage{nips15submit_e,times}
\usepackage{hyperref}
\usepackage{wrapfigure}
\usepackage{url}
\usepackage{amsmath}
\usepackage{amssymb}
\usepackage{bbm}
\usepackage{graphicx}
\usepackage{tabu}
\usepackage{subcaption}
\usepackage{babel}
\usepackage{subcaption}
\usepackage{booktabs}

\usepackage{rotating} 

\graphicspath{ {images/} }

\newcommand{\invisible}[1]{}

\newcommand{\bvb}{\mathbf b}

\newcommand{\ffix}{f_{\mathrm{fix}}}
\newcommand{\Lfix}{L_{\mathrm{fix}}}

\newcommand{\Lhung}{L_{\mathrm{hung}}}

\newcommand{\Lfirstk}{L_{\mathrm{firstk}}}

\newcommand{\fuf}{{\mathrm f}}

\title{End-to-end people detection in crowded scenes}

\author{
Russell Stewart$^1$ \qquad Mykhaylo Andriluka$^{1,2}$\\
$^1$Department of Computer Science, Stanford University, USA\\
$^2$Max Planck Institute for Informatics, Saarbr\"ucken, Germany\\
\texttt{\{stewartr,andriluk\}@stanford.edu}\\
}

\nipsfinalcopy 

\begin{document}

\maketitle

\begin{abstract}
  Current people detectors operate either by scanning an image in a sliding window fashion or by
  classifying a discrete set of proposals. We propose a model that is based on \textit{decoding} an
  image into a set of people detections. Our system takes an image as input and directly outputs a
  set of distinct detection hypotheses. Because we generate predictions jointly,
  common post-processing steps such as non-maximum suppression are unnecessary. We use a recurrent LSTM layer
  for sequence generation and train our model end-to-end with a new loss function that operates on sets of
  detections.
  We demonstrate the effectiveness of our approach on the challenging task of detecting people in crowded scenes.
\end{abstract}

\section{Introduction}

In this paper we propose a new architecture for detecting objects in images. We strive for an
end-to-end approach that accepts images as input and directly generates a set of object bounding boxes
as output. This task is challenging because it demands both distinguishing objects from
the background and correctly estimating the number of distinct objects and their locations.
Such an end-to-end approach capable of directly outputting predictions would be advantageous over methods that
first generate a set of bounding boxes, evaluate them with a classifier, and then perform some
form of merging or non-maximum suppression on an overcomplete set of detections. 

Generating a set of detections from an integrated process has an important advantage in that
multiple detections on the same object can be avoided by remembering the previously generated
output. To control this generation process, we use a recurrent neural network with LSTM units.  To
produce intermediate representations, we use expressive image features from GoogLeNet that are
further fine-tuned as part of our system. Our architecture can thus be seen as a ``decoding''
process that converts an intermediate representation of an image into a set of predicted
objects. The LSTM can be seen as a ``controller'' that propagates information between decoding steps
and controls the location of the next output (see Fig.~\ref{fig:model} for an overview). Importantly, our trainable end-to-end system allows
joint tuning of all components via back-propagation.

One of the key limitations of merging and non-maximum suppression utilized in
\cite{girshick2014rcnn,sermanet-iclr-14} is that these methods typically don't have access to image
information, and instead must perform inference solely based on properties of bounding boxes
(e.g. distance and overlap). This usually works for isolated objects, but often fails when object
instances overlap. In the case of overlapping instances, image information is necessary to decide
where to place boxes and how many of them to output. As a workaround, several approaches proposed
specialized solutions that specifically address pre-defined constellations of objects (e.g. pairs of
pedestrians) \cite{Farhadi:2011:RVP,Tang:2012:DTO}. Here, we propose a generic architecture that
does not require a specialized definition of object constellations, is not limited to pairs of
objects, and is fully trainable.

\begin{figure}
    \begin{subfigure}{0.31\textwidth}
        \includegraphics[width=\linewidth]{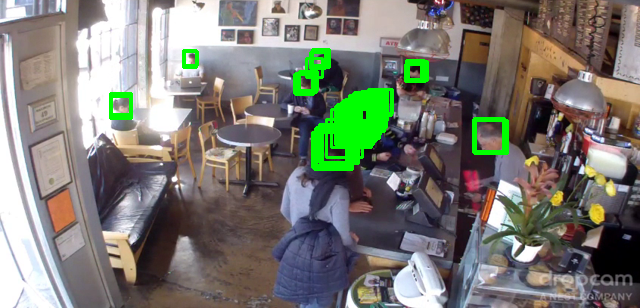}
        \caption{OverFeat output} \label{fig:teaser1}
    \end{subfigure}
    \hspace*{\fill} 
    \begin{subfigure}{0.31\textwidth}
        \includegraphics[width=\linewidth]{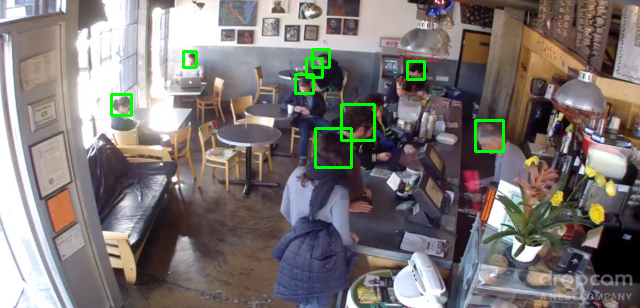}
        \caption{OverFeat final predictions} \label{fig:teaser2}
    \end{subfigure}
    \hspace*{\fill} 
    \begin{subfigure}{0.31\textwidth}
        \includegraphics[width=\linewidth]{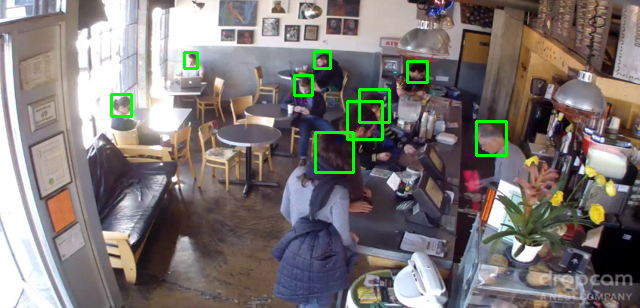}
        \caption{Our predictions} \label{fig:teaser3}
    \end{subfigure}
    \caption{Initial over-complete set of detections of OverFeat (a) and output of post-processing
      (b). Note the failure to detect the third person in the center. Detection results obtained
      with our \mbox{method (c)}.} \label{fig:1}
\end{figure}

We specifically focus on the task of people detection as an important example of this
problem. In crowded scenes such as the one shown in Fig.~\ref{fig:1}, multiple people often occur in close proximity, making it particularly
challenging to distinguish between nearby individuals. 

The key contribution of this paper is a trainable, end-to-end approach that jointly predicts the 
objects in an image. This lies in contrast to existing methods
that treat prediction or classification of each bonding box as an independent problem and
require post-processing on the set of detections. We demonstrate that our approach is superior to existing architectures on a
challenging dataset of crowded scenes with large numbers of people. A technical contribution of this
paper is a novel loss function for sets of objects that combines elements of localization and
detection. Another technical contribution is to show that a chain of LSTM units can be successfully
utilized to decode image content into a coherent real-valued output of variable length. We envision
this technique to be valuable in other structured computer vision prediction tasks such as
multi-person tracking and articulated pose estimation of multiple people.

\subsection{Related work}

Detection of multiple objects in the presence of occlusions has been a
notorious problem in computer vision. Early work employed a codebook of local features and Hough
voting \cite{Leibe:2005:PDC,Barinova:2010:ODM}, but still required complex tuning and multi-stage pipelines. Importantly, these
models utilized weak representations based on local features that are outperformed by modern
deep representations. 

To overcome the difficulties of predicting multiple objects in close proximity, several attempts have
been made to jointly predict constellations of objects
\cite{Farhadi:2011:RVP,Tang:2012:DTO,Ouyang:2013:SPD}. Our work is more general, as we do not
explicitly define these groups, and instead let the model learn any features that are necessary for finding occluded instances.

Currently, the best performing object detectors operate either by densely scanning the image in a sliding
window fashion \cite{sermanet-iclr-14,girshick2014rcnn,Zhang2015Cvpr}, or by using a proposal
mechanism such as \cite{Uijlings13,Szegedy:2014:SHQ}, and leveraging CNNs to classify a sparsified set of
proposals \cite{girshick2014rcnn}. These approaches work well for images with few object
instances that do not overlap, but often fail in the presence of strong
occlusions. For example, MultiBox \cite{Szegedy:2014:SHQ} learns class independent proposals that are subsequently
classified with CNN. Like MultiBox, we propose a set of bounding boxes from images, but these proposals
directly correspond to object instances and do not require post-processing. The MultiBox outputs are necessarily 
sparse, whereas our system is able to generate predictions in arbitrarily close proximity.

Our approach is related to the OverFeat model \cite{sermanet-iclr-14}. We rely on a regression module
to generate boxes from a CNN encoding. However, in our case boxes are generated as part of an integrated
process, and not independently as in OverFeat. As a result, each output box corresponds directly to an object detected in the image, and we do not require merging or non maximum suppression.
Another important advantage of our approach is that it outputs a confidence corresponding to each output that is trained end-to-end.
In the case of OverFeat, an end-to-end trained confidence prediction is not available, as the output is the result of a heuristic merging procedure.



Our work is related to recent neural network models for predicting sequences \cite{karpathy15cvpr,Sutskever:2014:STS}. As in
\cite{Sutskever:2014:STS}, we rely on an LSTM to predict variable length outputs. 
Unlike in language generation, detection requires that a system reason over a 2D output space, which lacks a natural linear ordering.
MultiBox \cite{Szegedy:2014:SHQ} addresses this challenge by introducing a loss function that allows unordered predictions to be permuted to match
ground-truth instances during training.
We build on this contribution by leveraging the capacity of our recurrent decoder to make joint predictions in sequence. 
In addition to computing an optimal matching of predictions to ground-truth, our loss function encourages the model to make predictions
in order of descending confidence. Suitable loss functions have previously been proposed in
structured speech recognition and natural language processing \cite{Graves:2006:CTC}. Here we propose such a loss
function for object detection. 

\section{Model}
\subsection{Overview}
\label{sec:modeldef}

\begin{figure}
\centering\includegraphics[width=\textwidth]{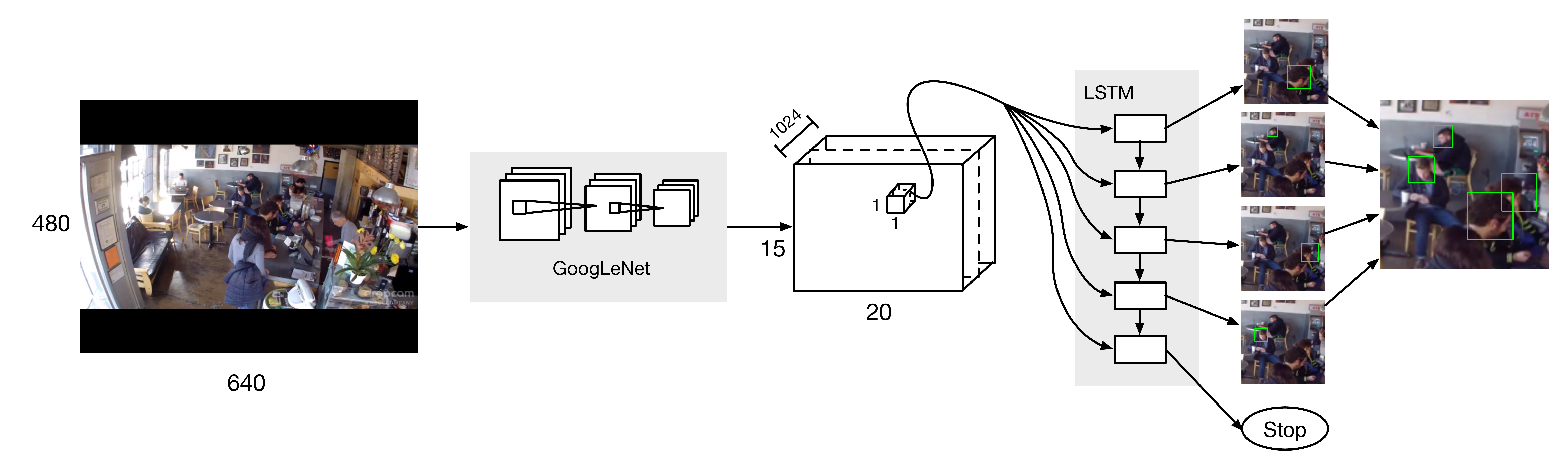}
\caption{Our system first encodes an image into a block of high level features. An LSTM then acts as a controller, decoding this information into a set of detections.}
  \label{fig:model}
\end{figure}

Deep convolutional architectures such as \cite{Krizhevsky:2012:ICD,Szegedy:2014:GDC} construct image
representations that are effective for a variety of tasks. These architectures have been leveraged
for detection, albeit primarily by adapting them into a classification or regression framework. 
Deep
representations have sufficient power to jointly encode the appearance of multiple
instances, but one must augment them with a component for multiple instance prediction to realize this potential. In this paper, we consider
recurrent neural networks (RNN), and in particular LSTM units \cite{Hochreiter:1997:LSTM} as a candidate for such a component.
The key properties that make the combination of deep CNN's with RNN-based decoders appealing are (1)
the ability to directly tap into powerful deep convolutional representations and (2) the ability to
generate coherent sets of predictions of variable length. These properties have been leveraged successfully in \cite{karpathy15cvpr}
to generate image captions, and in \cite{Sutskever:2014:STS} for machine translation. The ability to generate coherent sets is particularly important
in our case because our system needs to remember previously generated predictions and avoid multiple predictions of the same target.


We construct a model that first encodes an image into high level descriptors via a convolutional
architecture (e.g.~\cite{Szegedy:2014:GDC}), and then decodes that representation into a set of
bounding boxes. As a core machinery for predicting variable length output, we build on a recurring
network of LSTM units. An overview of our model is shown on Fig.~\ref{fig:model}.  We
transform each image into a grid of 1024 dimensional feature descriptors at strided regions
throughout the image. The 1024 dimensional vector summarizes the contents of the region and carries
rich information regarding the positions of objects. The LSTM draws from this
information source and acts as a controller in the decoding of a region.
At each step, the LSTM outputs a new bounding box and a corresponding confidence that a previously undetected
person will be found at that location. Boxes are encouraged to be produced in order of descending
confidence. When the LSTM is unable to find another box in the region with a confidence above
a prespecified threshold, a stop symbol is produced. The sequence of outputs is collected and presented as a
final description of all object instances in the region.


\subsection{Loss function} \label{loss_function}
\label{sec:lossfunc}

The architecture introduced in Sec.~\ref{sec:modeldef} predicts a set of candidate bounding boxes
along with a confidence score corresponding to each box. Hypotheses are generated in sequence and
later predictions depend on previous ones via the memory states of the LSTM. At each recurrence, the
LSTM outputs an object bounding box $\bvb=\{\bvb_{pos},b_c\}$, where
$\bvb_{pos}=(b_x, b_y, b_w, b_h) \in \mathbb{R}^4$ is a relative position, width and height of the
bounding box, and $b_c\in [0,1]$ is a real-valued confidence. Confidence values lower than a pre-specified
threshold (e.g. 0.5) will be interpreted as a stop symbol at test time. Higher values of the bounding box
confidence $b_c$ should indicate that the box is more likely to correspond to a true positive. We
denote the corresponding set of ground truth bounding boxes as $G = \{\bvb^i|i=1,\ldots, M\}$,
and the set of candidate bounding boxes generated by the model as $C = \{\tilde\bvb^j|j=1,\ldots, N\}$.
In the following we introduce a loss function suitable for guiding the learning process towards the desired output.

\begin{wrapfigure}{r}{0.4\textwidth}
\centering\includegraphics[width=3cm]{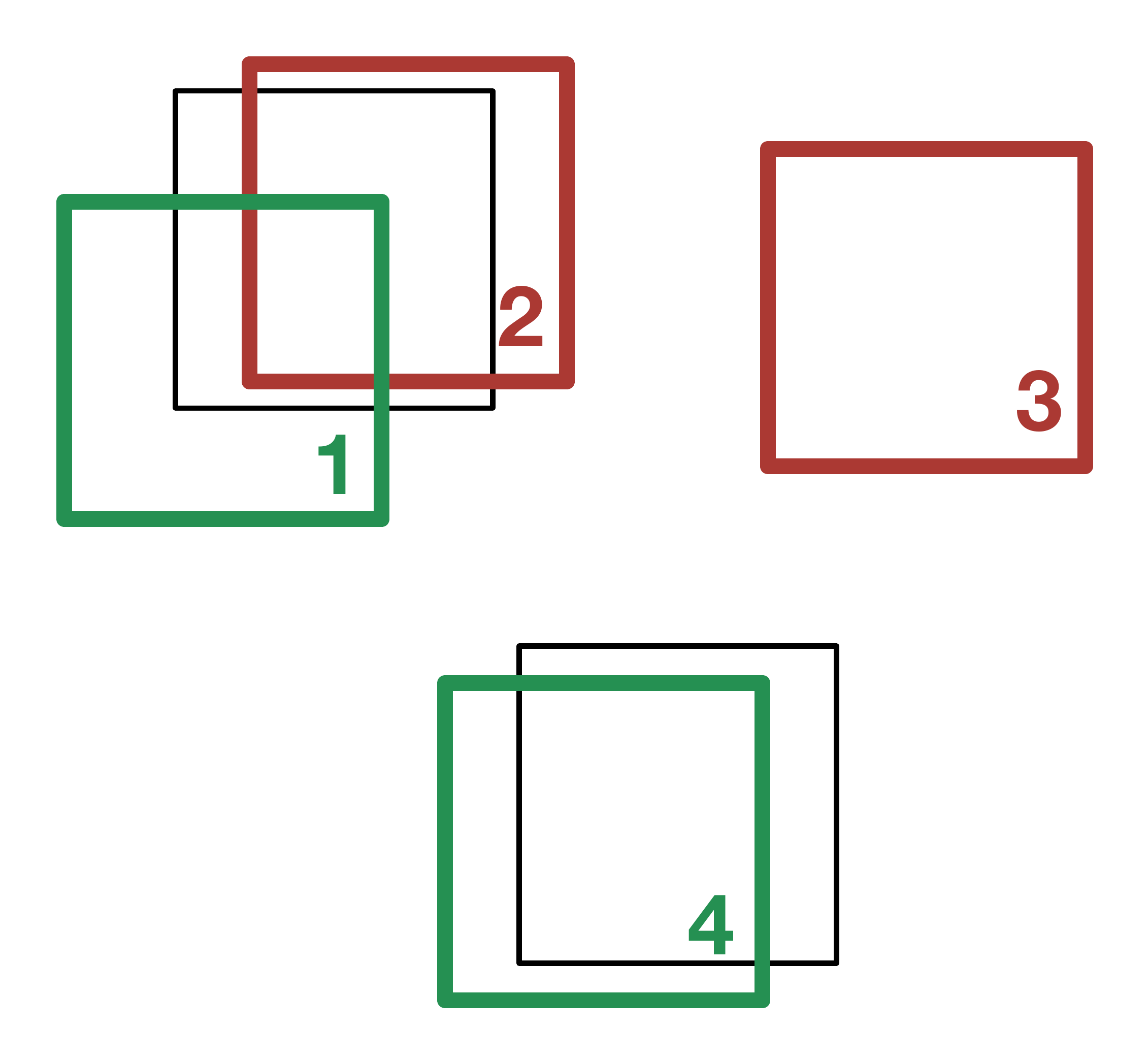}
\caption{Illustration of the matching of ground-truth instances (black) to accepted (green)
    and rejected (red) candidates. Matching should respect both precedence (1 vs 2) and localization (4 vs 3).}
  \label{fig:matching}
\end{wrapfigure}

Consider the example in Fig.~\ref{fig:matching}, which schematically shows a detector 
with four generated hypotheses, each numbered by
its prediction step, which we denote as rank. Note 
the typical detection mistakes such
as false positives (hypothesis 3), imprecise localizations (hypothesis 1), and multiple predictions of
the same ground-truth instance (hypotheses 1 and 2). Different mistakes require different kinds of feedback.
In the case of hypothesis 1, the box location must be fine-tuned.  Conversely, hypothesis 3 is a false positive, and the model should instead abandon
the prediction by assigning a low confidence score. Hypothesis 2 is a second
prediction on the target already reported by hypothesis 1, and should be abandoned as well.
To capture these relationships, we introduce a matching algorithm that assigns a unique candidate hypothesis to each ground-truth.
The algorithm returns an injective function
$\fuf: G \rightarrow C$ ,
i.e.~$\fuf(i)$ is the index of candidate hypothesis assigned to ground-truth hypothesis $i$.

Given $\fuf$, we define a loss function on pairs of sets $G$ and $C$ as

\begin{align}
    L(G, C, f) = \alpha \sum_{i = 1}^{|G|}l_{pos}(\bvb^i_{pos}, \tilde\bvb^{f(i)}_{pos}) + \sum_{j=1}^{|C|} l_{c}(\tilde\bvb^j_c, y_j)
\label{eq:loss}
\end{align}

\noindent where $l_{pos} = \|\bvb^i_{pos} - \tilde\bvb^{f(i)}_{pos}\|_1$ is a displacement between
the position of ground-truth and candidate hypotheses, and $l_{c}$ is a cross-entropy loss on a
candidate's confidence that it would be matched to a ground-truth. The label for this cross-entropy loss is provided by $y_j$.
It is defined from the matching function as
$y_j=\mathbbm{1}\{f^{-1}(j) \neq \varnothing\}$. $\alpha$ is a term trading off between confidence errors and localization errors. We set $\alpha = 0.03$ with cross validation. Note that for a fixed matching, we can update the network
by backpropagating the gradient of this loss function.

As an na\"{\i}ve baseline, we consider a simple matching strategy based on the fixed ordering of the
ground-truth bounding boxes. We sort ground-truth
boxes by image position from top to bottom and from left to right. This fixed order matching
sequentially assigns candidates to the sorted ground-truth. We refer to this matching function as
``fixed order'' matching, denoting it as $\ffix$, and the corresponding loss function as $\Lfix$.

\paragraph{Hungarian loss:} The limitation of the fixed order matching is that it might
incorrectly assign candidate hypotheses to ground-truth instances when the decoding process produces false positives
or false negatives. This issue persists for any specific ordering chosen by $\ffix$. We
thus explore loss functions that consider all possible one-to-one assignments between elements in $C$ and $G$.


Recall that one of the principled objectives of our model is to output a coherent sequence of
predictions on multiple objects. We define the stopping criterion for the generation process to be
when a prediction score falls below a specified threshold. For such a score threshold to make sense,
we must encourage the model to generate correct hypotheses early in the sequence,
and to avoid generating low-confidence predictions before high-confidence ones. Therefore, when two
hypotheses both significantly overlap the same ground-truth (e.g. hypotheses 1 and 2 in
Fig.~\ref{fig:matching}), we prefer to match the hypothesis that appears earlier in the
predicted sequence.

To formalize this notion, we introduce the following comparison function between hypotheses and
ground-truth: 
\begin{align}
    \Delta(\bvb_i, \tilde\bvb_j) = (o_{ij},r_i,d_{ij})
\label{eq:comparetuple}
\end{align}

The function $\Delta: G \times C \rightarrow \mathbb{N} \times \mathbb{N} \times \mathbb{R}$ returns a
tuple where $d_{ij}$ is the $L_1$ distance between bounding box locations, $r_j$ is the rank or index of
$\tilde\bvb_j$ in the prediction sequence output by the LSTM, and $o_{ij} \in \{0, 1\}$ is a variable penalizing hypotheses that do not sufficiently
overlap a ground-truth instance.
Here, the overlapping criterion requires that a candidate's center lie within the extent of the
ground-truth bounding box. The $o_{ij}$ variable makes an explicit distinction between localization and
detection 
errors.
We define a lexicographic ordering on tuples produced by $\Delta$. That is, when evaluating which of two hypotheses will be assigned to a ground-truth,
overlap is paramount, followed by rank and then fine-grained localization.

Given the definition of the comparison function $\Delta$ in Eq.\ref{eq:comparetuple}, we find the minimal cost bipartite
matching between $C$ and $G$ in polynomial time via the Hungarian algorithm. Note that the Hungarian algorithm is
applicable to any graph with edge weights that have well-defined addition and pairwise comparison
operations. To that end, we define $(+)$ as element-wise addition and $(<)$ as lexicographic
comparison. For the example in Fig.~\ref{fig:matching}, correctly matching hypotheses $1$ and $4$ would
cost $(0, 5, 0.4)$, whereas matching $1$ and $3$ would cost $(1, 4, 2.3)$, and matching $2$ and $4$
would cost $(0, 6, 0.2)$. Note how the first term, used for detecting overlap, properly handles the case where a hypothesis
has low rank, but is too far from the ground-truth to be a sensible match (as is the case for hypothesis 3 in
Fig.~\ref{fig:matching}). We refer to the corresponding loss for this matching as the Hungarian loss and denote is as
$\Lhung$.

We also consider a simplified version of $\Lhung$ where only the top $k = |G|$ ranked predictions
from $C$ are considered for matching. Note that this is equivalent to removing or zeroing out the pairwise matching terms
$o_{ij}$ in Eq.~\ref{eq:comparetuple}. We denote this loss as $\Lfirstk$. We
experimentally compare $\Lfix$, $\Lfirstk$, and $\Lhung$ in Sec.~\ref{sec:expvariants}, showing that
$\Lhung$ leads to best results.

\section{Implementation details}

We constructed our model to encode an image into a 15x20 grid of 1024-dimensional top level GoogLeNet features. Each cell in the grid has a receptive field of size 139x139,
and is trained to produce a set of distinct bounding boxes in the center 64x64 region. The 64x64 size was chosen to be large enough to capture challenging local 
occlusion interactions. Larger regions may also be used, but provide little additional on our scenes, where few occlusion interactions span that scale.
300 distinct LSTM controllers are run in parallel, one for each 1x1x1024 cell of the grid. 

Our LSTM units have 250 memory states, no bias terms, and no output nonlinearities. At each step, we concatenate the GoogLeNet features with the output of the previous LSTM unit, and feed the result into
the next LSTM unit.
We have produced comparable results by only feeding the image into the first LSTM unit, indicating that multiple presentations of the image may not be necessary.
Producing each region of the full 480x640 image
in parallel gives an efficient batching of the decoding process. 

Our model must learn to regress on bounding box locations through the LSTM decoder. During training, the decoder outputs an overcomplete set of bounding boxes,
each with a corresponding confidence. For simplicity and batching efficiency, the cardinality of the overcomplete set is fixed, regardless of the number of ground-truth
boxes. This trains the LSTM to output high confidence scores and correct localizations for boxes corresponding to the ground
truth, and low confidence scores elsewhere. Because early outputs are preferred during matching, the model learns to output
high confidence, easy boxes first. 
In our dataset, few regions
have more than 4 instances, and we limit the overcomplete set to 5 predictions. Larger numbers of predictions neither improved
nor degraded performance. 

After sequence generation, ground-truth instances are matched to predictions using the bipartite matching function described in \ref{sec:lossfunc},
which favors candidates that are output earlier and closer to the ground-truth targets. With an optimal matching determined, we backpropagate the
Hungarian loss described in equation \ref{eq:loss} through the full network.

\paragraph{Model training:} We use the Caffe open source deep learning framework \cite{jia2014caffe} for training and evaluation.
The decoder portion of our model is a custom LSTM implementation derived from NLPCaffe \cite{Stewart:2015:NLPCaffe}. We train with
learning rate $\epsilon = 0.2$ and momentum 0.5. Gradients are clipped to have
maximum 2-norm of 0.1 across the network.  We decreased the learning rate by a multiple of 0.8 every
100,000 iterations. Convergence is reached at 500,000 iterations.

Training proceeds on all subregions of one image at each iteration. Parallelism of the LSTM decoders across regions mitigates efficiency gains for larger batch sizes. All weights are tied between regions
and LSTM steps. However, we were surprised to find slight performance gains when using separate weights connecting
LSTM outputs to predicted candidates at each step. These weights remain tied across regions. Tying these weights reduced average precision (AP) from $.85$ to $.82$. All hyperparameter AP analysis is performed on the validation set.

\begin{figure}[t]
\centering
  \includegraphics[width=0.65\linewidth]{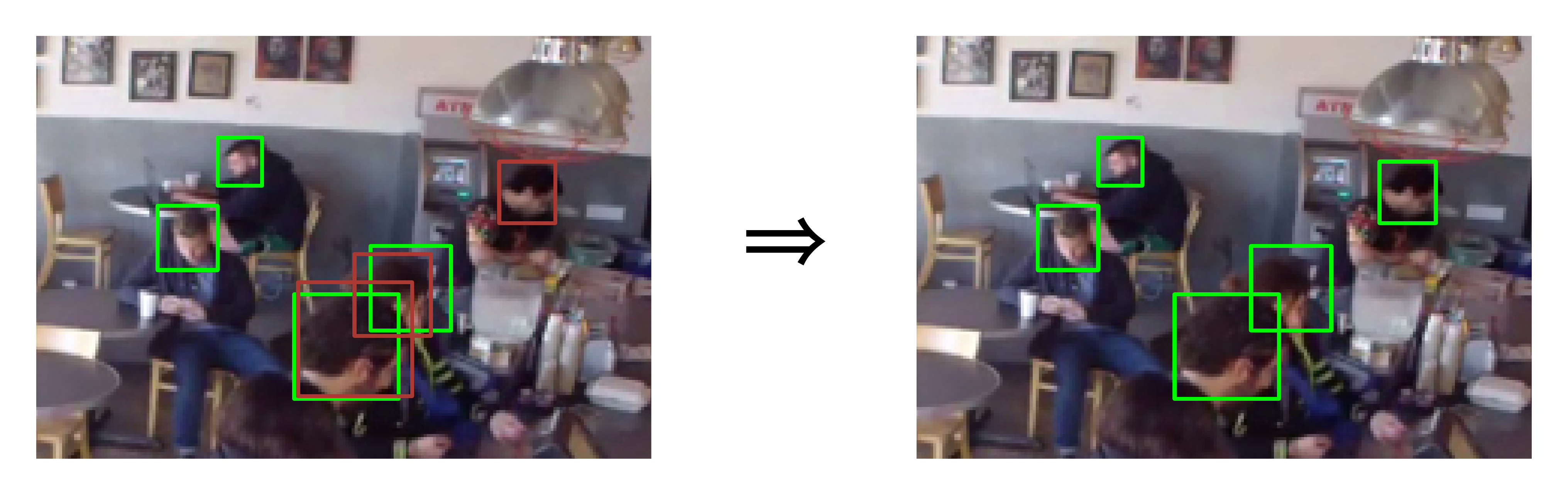}
  \caption{Example of stitching in a new region's predictions (red) with accepted predictions (green).}\label{fig:stitching}
\end{figure}


\paragraph{Initialization:} GoogLeNet weights are initialized to a snapshot after 10,000 iterations of training from our baseline OverFeat model (which was in turn initialized with Imagenet pretraining).
Fine tuning of GoogLeNet features to meet the new demands of the decoder is critical. Training without fine tuning GoogLeNet reduced AP by $.29$. 

All weights in the decoder are initialized from a uniform distribution in [-0.1, 0.1]. However, these initializations differ drastically from
our pretrained GoogLeNet, which has activations in the range [-80, 80]. To compensate for this mismatch, we use a scale layer to decrease GoogLeNet activations
by a factor of 100 before feeding them into the LSTM. Likewise, the initial standard deviation of the fully connected layers output is on the order of 0.3,
but bounding box pixel locations and sizes vary in [-64, 64]. Thus, we scale up the regression predictions by a factor of 100
before comparing them with ground-truth. Note that these modifications are the same as changing weight initializations only if one also introduces proportional learning rate multipliers.


\paragraph{Regularization:} We use dropout with probability 0.15 on the output of each LSTM. Removing dropout reduces AP by $.011$. Images are
jittered by up to 32 pixels in the vertical and horizontal directions, and are scaled at random by a factor in [0.9, 1.1]. We found it important
to remove $L_2$ regularization entirely. When applying the original $L_2$ regularization multiplier of 2e-4 to only the GoogLeNet section of our network,
we were unable to train. Multipliers as small as 1e-6 reduced AP by $.03$.

\paragraph{Stitching:} While we trained our algorithm to predict bounding boxes on 64x64 pixel regions, we apply our algorithm to full 480x640 images at test time. To this end, we generate predictions
from each region in a 15x20 grid of the image. We then use a stitching algorithm to recursively merge in predictions from successive cells on the grid.
Each iteration of the algorithm performs the process illustrated in Figure \ref{fig:stitching}. At a given iteration, let $A$ denote the current set of all
accepted bounding box predictions. We process a new region, evaluating the decoder until a stop symbol is produced and collect a set $C$ of newly proposed bounding boxes. 
Some of these new bounding boxes may correspond to previous predictions. Thus to avoid adding false positives, we
destroy any new boxes having nonzero intersection with accepted boxes, conditioned on the constraint that previously accepted boxes may destroy at most one new box. 

To this end, we define a bipartite matching problem related to that in section \ref{loss_function} with a pairwise loss term $\Delta': A
\times C -> \mathbb{N} \times \mathbb{R}$ given as $\Delta'(\bvb_i,\tilde\bvb_j) = (m_{ij},
d_{ij})$. Here, $m_{ij}$ states whether two boxes do not intersect, and $d_{ij}$ is a local
disambiguation term given by the $L_1$ distance between boxes.
Minimizing this matching cost will maximize the number of destroyed candidates. As before, we leverage the Hungarian algorithm to find a minimum cost matching in polynomial time. 
We examine each match pair, $(\bvb, \tilde\bvb)$, and add any candidate $\tilde\bvb$ that does not overlap with its match $\bvb$ to the set of accepted boxes.






\section{Experimental results}

\label{sec:expvariants}

\begin{figure}[t]\label{fig:some}
\begin{subfigure}{.6\textwidth}
  \centering
 \includegraphics[width=4.cm]{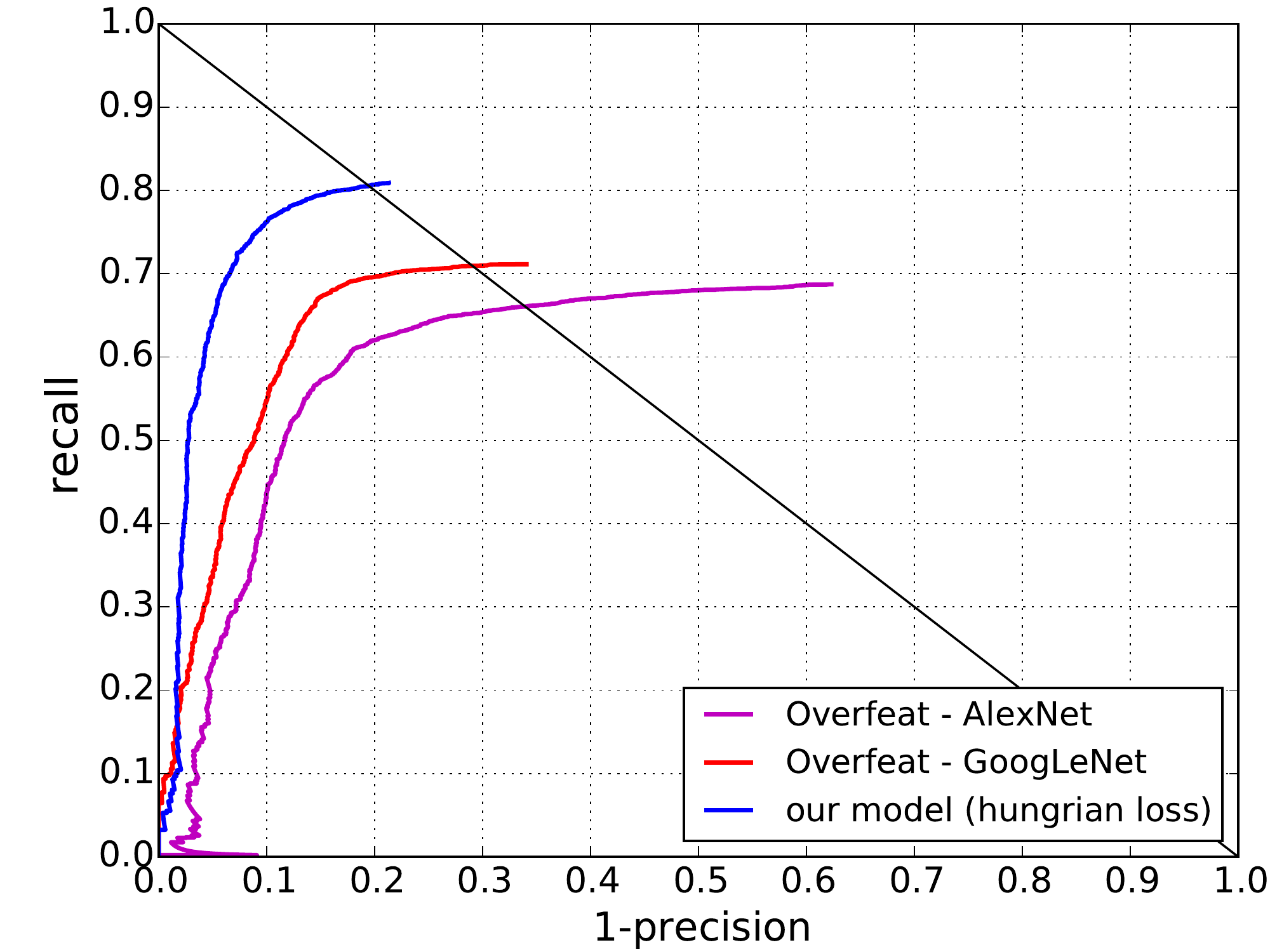}
 \includegraphics[width=4.cm]{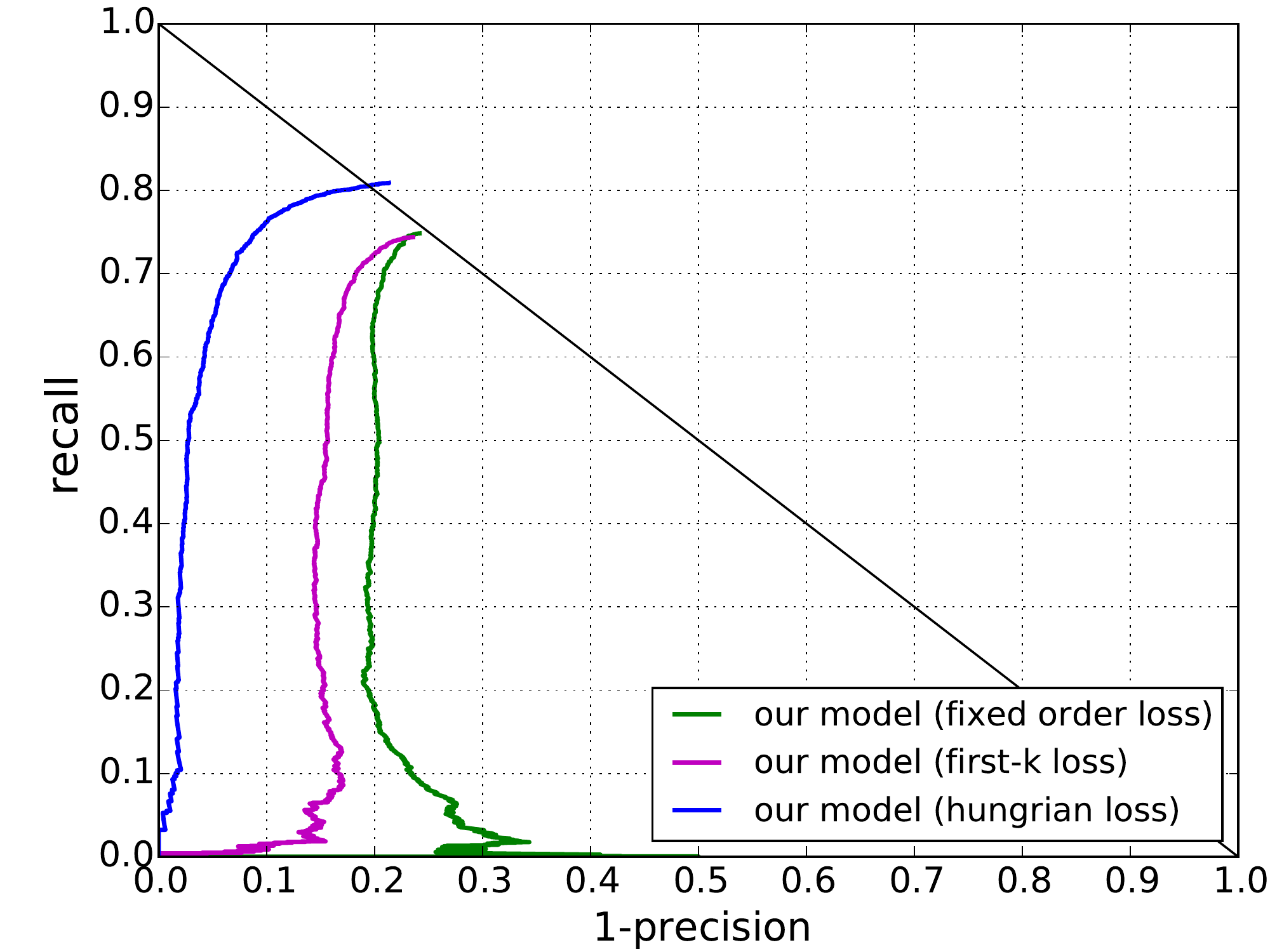}
\label{fig:compare}
\end{subfigure}%
\begin{subfigure}{.4\textwidth}

 \scriptsize
  \centering
  \begin{tabular}{@{} l ccc @{}}
    \toprule
    Method & AP   & EER  & COUNT \\
    \midrule
    Overfeat - AlexNet & 0.62 & 0.66 & 1.29 \\
    Overfeat - GoogLeNet & 0.67 & 0.71 & 1.05\\
    our model, $\Lfix$ & 0.60 & 0.75 & 0.80 \\
    our model, $\Lfirstk$ &  0.63 & 0.75 & \textbf{0.74} \\
    our model, $\Lhung$ & \textbf{0.78} & \textbf{0.81} & 0.76 \\
    \bottomrule
  \end{tabular}
  \vspace{0.5em}
  \label{tab:compare_others}

\end{subfigure}
\caption{Performance evaluation.}
\label{fig:rpc}
\end{figure}


\paragraph{Dataset and evaluation metrics:} To evaluate the performance of our approach, we collected a large dataset of
images from busy scenes using video footage available from public webcams. In total, we
collect $11917$ images with $91146$ labeled people. We extract images from video footage at a fixed
interval of $100$ seconds to ensure a large variation in images. We allocate $1000$
images for testing and validation, and leave the remaining images for training, making sure that no
temporal overlaps exist between training and test splits. The resulting training set contains
$82906$ instances. Test and validation sets contain $4922$ and $3318$ people instances
respectively. Images were labeled using Amazon Mechanical Turk by a handful of workers pre-selected
through their performance on an example task. We label each person's head to
avoid ambiguity in bounding box locations. The annotator labels
any person she is able to recognize, even if a substantial part of the person is not visible.
Images and annotations will be made available\footnote{\url{http://d2.mpi-inf.mpg.de/datasets}}.

Examples of collected images are shown in Fig.~\ref{fig:detections}, and in the video
included in the supplemental material. Images in our dataset include challenges such as people at small scales,
strong partial occlusions, and a large variability in clothing and appearance. Our evaluation uses the standard protocol defined in \cite{Everingham15}. A hypothesis is
considered correct if its intersection-over-union score with a ground-truth bounding box is larger
than $0.5$. We plot recall-precision curves and summarize results in each experiment with average
precision (AP) and equal error rate (EER) in Fig.~\ref{fig:rpc}. We also analyze how well each model predicts the total count
of people in an image. As in \cite{Lempitsky:2010:LTC}, we measure count error by computing the average absolute
difference between the number of predicted and ground-truth detections in test set images.
For each model, an optimal detection threshold is selected on the validation set,
and we report the results as COUNT in Fig.~\ref{fig:rpc}.


\paragraph{Baseline methods:} We experimented with R-CNN \cite{girshick2014rcnn} and OverFeat \cite{sermanet-iclr-14} models to
define a strong baseline in our comparison. We use the implementation of OverFeat for the Caffe
framework \cite{jia2014caffe} that has been made publicly available by \cite{Huval:2015:AEE}, and R-CNN provided
by \cite{girshick2014rcnn}. In the initial experiments, R-CNN did not appear suitable for our setting primarily
because 
the bottom-up proposals generated with selective search \cite{Uijlings13} achieved poor recall. We
therefore use OverFeat in the rest of the evaluation. The original version of OverFeat provided by
\cite{Huval:2015:AEE} relied on the image representation based on AlexNet \cite{Krizhevsky:2012:ICD}. We hence refer to
the original version as OverFeat-AlexNet. Since both OverFeat and our model are implemented in Caffe, we
were able to directly substitute the GoogLeNet architecture into the OverFeat model. We denote the new
model as OverFeat-GoogLeNet. The comparison of the two OverFeat variants is shown in
Fig.~\ref{fig:rpc}. We observe that Overfeat-GoogLeNet performs significantly better than OverFeat-AlexNet.


\paragraph{Performance evaluation:} Note that the image representation used in our model and in
OverFeat is exactly the same. Both are implemented using the same code, parameters, filter
dimensions, and number of filters. This gives us the interesting possibility of directly comparing
the hypothesis generating components of both models. In the case of OverFeat
\cite{sermanet-iclr-14}, this component corresponds to a bounding box regression from each cell
followed by a round of non-maximum suppression. In our model this component corresponds to decoding
with an LSTM layer that produces a variable length output. The performance of our best model is
shown Fig.~\ref{fig:rpc} and compared to both versions of
OverFeat. 
Our approach delivers a substantial improvement over OverFeat, improving recall from $71\%$ to
$81\%$. We also achieve considerable improvement in AP ($0.78$ for our model vs. $0.67$ for
OverFeat-GoogLeNet), and people counting error ($0.76$ vs. $1.05$). Fig.~\ref{fig:detections} shows
several examples of detections obtained by our model and OverFeat-GoogLeNet. The green arrows
highlight cases where our model can detect people even in the presence of strong occlusions.
Examples of typical failure cases are indicated by red arrows in Fig.~\ref{fig:mistakes}.


\bgroup
\tabcolsep 1.0pt
\renewcommand{\arraystretch}{0.25}
\newlength{\teaserwidth}
\setlength{\teaserwidth}{1.85cm}
\begin{figure}
\center
\begin{tabular}{ccccc}
\begin{sideways} \tiny{OverFeat-GoogLeNet}  \end{sideways} & \hspace{-0.07in}
\includegraphics[height=\teaserwidth]{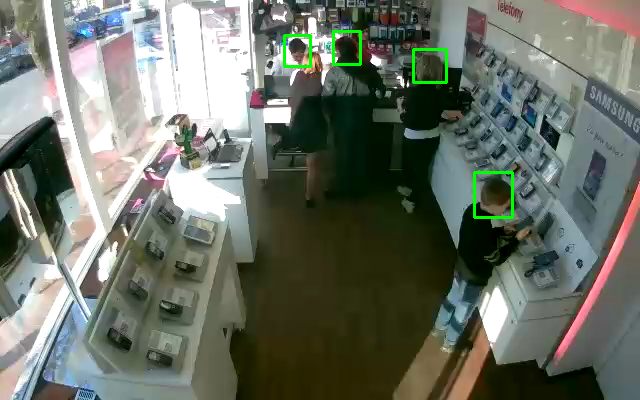}&
\includegraphics[height=\teaserwidth]{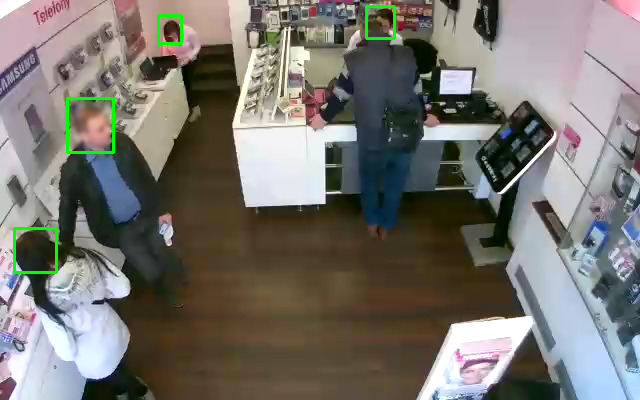}&
\includegraphics[height=\teaserwidth]{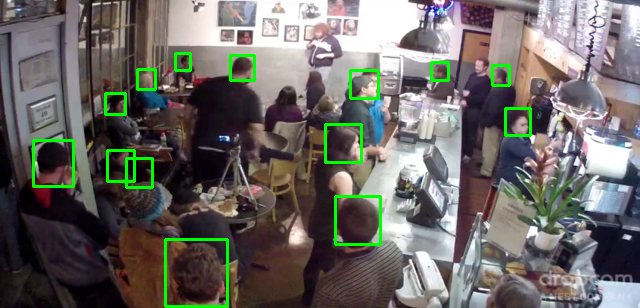}&
\includegraphics[height=\teaserwidth]{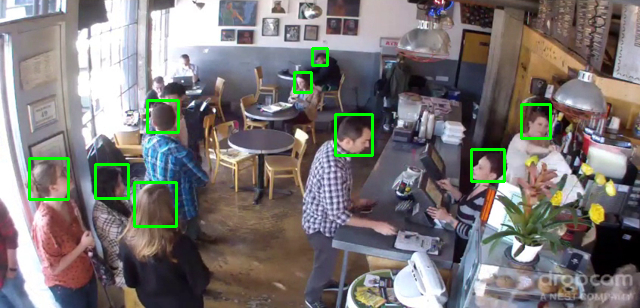}\\[1pt]
\begin{sideways} ~~ \tiny{our approach}\end{sideways} & \hspace{-0.07in}
\includegraphics[height=\teaserwidth]{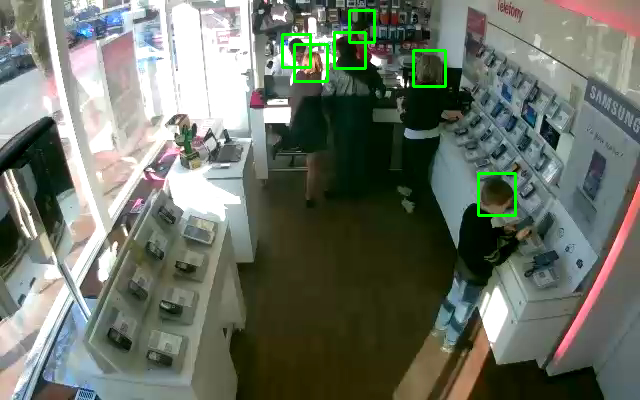}&
\includegraphics[height=\teaserwidth]{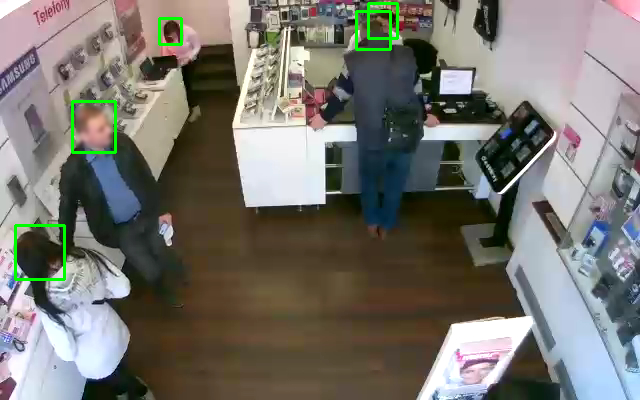}&
\includegraphics[height=\teaserwidth]{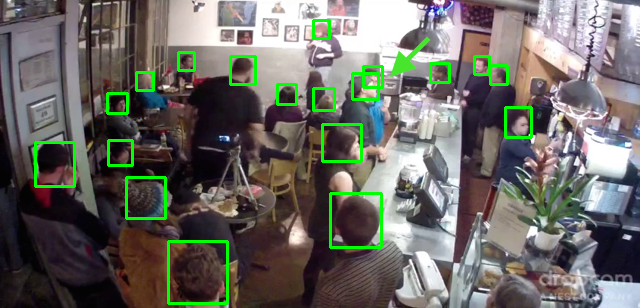}&
\includegraphics[height=\teaserwidth]{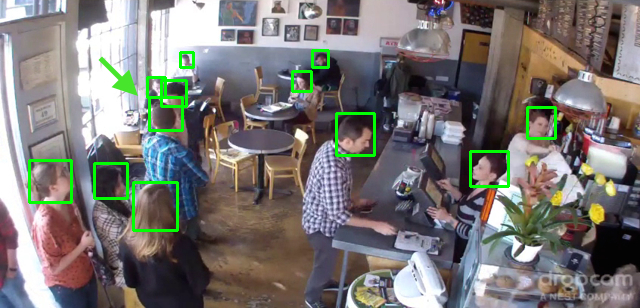}\\[1pt]
 \end{tabular} 
\caption{Example detection results obtained with OverFeat-GoogLeNet (top row) and our approach (bottom
  row). We show each model's output at $90\%$ precision. See the text for a description and the
  supplemental material for a video\protect\footnote{} of our model's output. }
\label{fig:detections}
\vspace{-5pt}
\end{figure}
\egroup
\footnotetext{\url{http://www.youtube.com/watch?v=QeWl0h3kQ24}}

\bgroup
\tabcolsep 1.0pt
\renewcommand{\arraystretch}{0.25}
\newlength{\mistakeswidth}
\setlength{\mistakeswidth}{1.9cm}
\begin{figure}
\center
 \begin{tabular}{cc}
\includegraphics[height=\mistakeswidth]{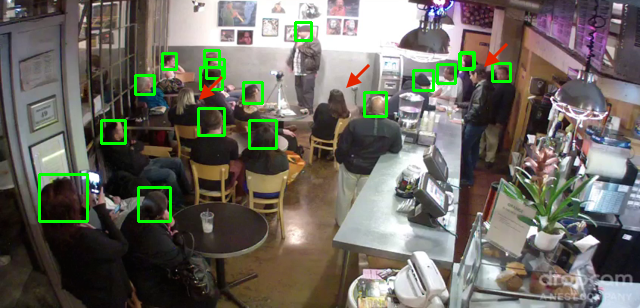} &
\includegraphics[height=\mistakeswidth]{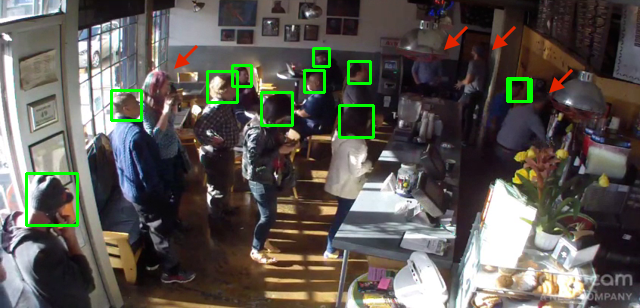}\\[1pt]
 \end{tabular} 
 \caption{Example failure cases of our method.}
\label{fig:mistakes}
\vspace{-5pt}
\end{figure}
\egroup

\paragraph{Comparison of loss functions.} We now evaluate the loss functions introduced in
Sec.~\ref{sec:lossfunc}. The model trained with $\Lfix$ achieves only $0.60$ AP. This suggests that
allowing the LSTM to output detections from easy to hard during training, rather than in some fixed spatial ordering,
was essential for performance. To explore the importance of overlap terms in our loss function, we
evaluate the $\Lfirstk$ loss, which matches the $k$ ground-truth instances in each region to the first $k$ output
predictions. We observe that $\Lfirstk$ outperforms $\Lfix$ at test time by allowing
permutations of LSTM outputs during training. However, we found that $\Lfirstk$ struggled to
attach confidences to specific box locations. With $\Lfirstk$, early
confidence predictions are often too high, and late predictions too low. It appears that instead of
learning the probability that the corresponding box is correct, the model learns on the $i^{th}$
recurrent step to predict the confidence that there are at least $i$
people in a region. These confidences are inappropriate for detection thresholding, and underscore
the importance of including the overlap terms, $o_{ij}$, in our matching function.
Precision recall curves for each loss function are shown in Fig.~\ref{fig:rpc}.

\section{Conclusion}

In this paper, we introduced a new method for object detection and demonstrated its performance on a
large dataset of images of crowded scenes. Our system addresses the challenge of detecting
multiple partially occluded instances by decoding a variable number of outputs from rich intermediate
representations of an image. To teach our model to produce coherent sets of predictions, we defined a loss function suitable for
training our system end-to-end. We envision that this approach
may also prove effective in other prediction tasks with structured outputs, such as people tracking
and articulated pose estimation.

\paragraph{Acknowledgements.} This work has been supported by the Max Planck Center for Visual
Computing and Communication. The authors would like to thank NVIDIA Corporation for providing a K40 GPU and
Angelcam\footnote{\url{www.angelcam.com}} for providing some of the images used in this work. The authors
would also like to thank Will Song and Brody Huval for helpful discussions.




\bibliographystyle{plain}
\bibliography{short,biblio}

\end{document}